\documentclass[10pt,lettersize,journal]{IEEEtran}

\IEEEoverridecommandlockouts                          

\usepackage{amsmath, amssymb}
\usepackage{algorithm}
\usepackage{algorithmic}
\usepackage{graphicx}
\usepackage{titlesec}
\usepackage{amsmath,amsfonts}
\usepackage{array}
\usepackage{caption}
\usepackage[caption=false,font=normalsize,labelfont=sf,textfont=sf]{subfig}
\usepackage{textcomp}
\usepackage{stfloats}
\usepackage{url}
\usepackage{verbatim}
\usepackage[margin=1in]{geometry} 
\usepackage{tabularx} 
\usepackage{booktabs}
\usepackage{makecell}
\usepackage{subcaption}
\usepackage{flushend}
\usepackage{mathtools}
\usepackage{hyperref}
\usepackage{tabularray}
\usepackage{pbox}
\usepackage{float}
\usepackage[noadjust]{cite}
\newcommand{\gap}{\vspace{0.1cm}}
\newcommand{\newsec}[1]{\gap \noindent {\bf #1}}

\title{\LARGE \bf
Bootstrapping Reinforcement Learning with Sub-optimal Policies for Autonomous Driving}
 
\author{Zhihao Zhang$^{1}$ Chengyang Peng$^{2}$ Ekim Yurtsever$^{3}$ and  Keith A. Redmill$^{4}$
\thanks{*This work is funded in part by Carnegie Mellon University’s Safety21 National University Transportation Center, which is sponsored by the US Department of Transportation under grants 69A3552344811/69A3552348316.}
\thanks{Some icons and graphics in this work were sourced from Vecteezy.com under the Free License.}
\thanks{$^{1}$Zhihao Zhang is a student of Electrical and Computer Engineering,
        The Ohio State University, 
        {\tt\small zhang.11606@osu.edu}}%
\thanks{$^{2}$Chenyang Peng is student of Mechanical Engineering,
        The Ohio State University, 
        {\tt\small peng.947@osu.edu}}%
\thanks{$^{3}$Dr. Ekim Yurtsever is a research assistant in The Center for Automotive Research,
        The Ohio State University, 
        {\tt\small yurtsever.2@osu.edu}}%
\thanks{$^{4}$Dr. Keith A. Redmill is with the Department of Electrical and Computer Engineering, The Ohio State University,
        {\tt\small redmill.1@osu.edu}}%
}

\begin{document}

\maketitle

\begin{abstract}

Automated vehicle control using reinforcement learning (RL) has attracted significant attention due to its potential to learn driving policies through environment interaction. However, RL agents often face training challenges in sample efficiency and effective exploration, making it difficult to discover an optimal driving strategy. To address these issues, we propose guiding the RL driving agent with a demonstration policy that need not be a highly optimized or expert-level controller. Specifically, we integrate a rule-based lane change controller with the Soft Actor Critic (SAC) algorithm to enhance exploration and learning efficiency. Our approach demonstrates improved driving performance and can be extended to other driving scenarios that can similarly benefit from demonstration-based guidance.
\end{abstract}

\vspace{-0.2in}
\section{INTRODUCTION}
Designing autonomous driving systems requires a modular approach that integrates key components such as vehicle sensing, perception, localization, scene representation, path planning, decision-making, and vehicle control~\cite {yurtsever2020survey}. Our study focuses on decision-making in driving scenarios. Conventional approaches to this problem often rely on model-based or rule-based controllers~\cite{bevly2016lane,hatipoglu2003automated,chandler1958traffic,gipps1981behavioural,treiber2000congested,kesting2007general,yurtsever2020survey}. Rule-based controllers, in particular, depend on human expertise to define detailed safety measures and scenario-specific rules, ensuring stability and avoiding situations where the vehicle might become stuck. Supervised learning methods, such as behavior cloning, aim to imitate expert driving behavior by training on large datasets of human driving~\cite{han2019driving,bojarski2016end,bojarski2017explaining,codevilla2018end,hawke2020urban}. Although capable of learning end-to-end policies, their performance is highly dependent on the quality and quantity of the expert driving data. Furthermore, their ability to generalize to unseen conditions is often limited, making them less reliable in complex or unforeseen scenarios.

In contrast, reinforcement learning (RL) represents a self-supervised approach, learning control policies through interactions with the environment. Deep Reinforcement Learning (DRL), leveraging neural networks, has demonstrated remarkable efficiency in mapping state-action transitions~\cite{sutton1998reinforcement}. Algorithms such as Proximal Policy Optimization (PPO)~\cite{schulman2017proximal} and SAC~\cite{haarnoja2018soft} have been developed to handle high-dimensional representations of the environment, enabling more robust solutions for decision-making tasks.

One of the primary challenges in RL is achieving sufficient exploration of the environment while maintaining sample efficiency. Despite methods such as maximum entropy DRL algorithms that promote continuous exploration during training, agents still struggle to identify optimal policies in complex environments that require long-term planning and effective handling of delayed rewards~\cite{haarnoja2018soft,schulman2017proximal}.

In our previous study, we found that traditional online RL driving agents can encounter dense traffic bottleneck situations where multiple steps of undesirable rewards discourage exploration and prevent the discovery of long-term beneficial solutions~\cite {zhang2025extensive}. Consequently, this limited exploration can cause the agents to converge on suboptimal policies such as overly conservative driving strategies. To address this issue, we propose a novel approach that incorporates a demonstration controller to guide RL agents. This auxiliary demonstration controller need not be carefully designed to ensure optimal performance, but it can help agents to overcome exploration barriers and converge on an optimal driving policy by providing plausible and human-like behavior that is less obvious and more difficult to discover. Thus, a good suboptimal demonstration policy should encourage the learning of less obvious behaviors while still being general and simple in design.

\begin{figure}[!t]
\centering
\includegraphics[width=0.5\textwidth]{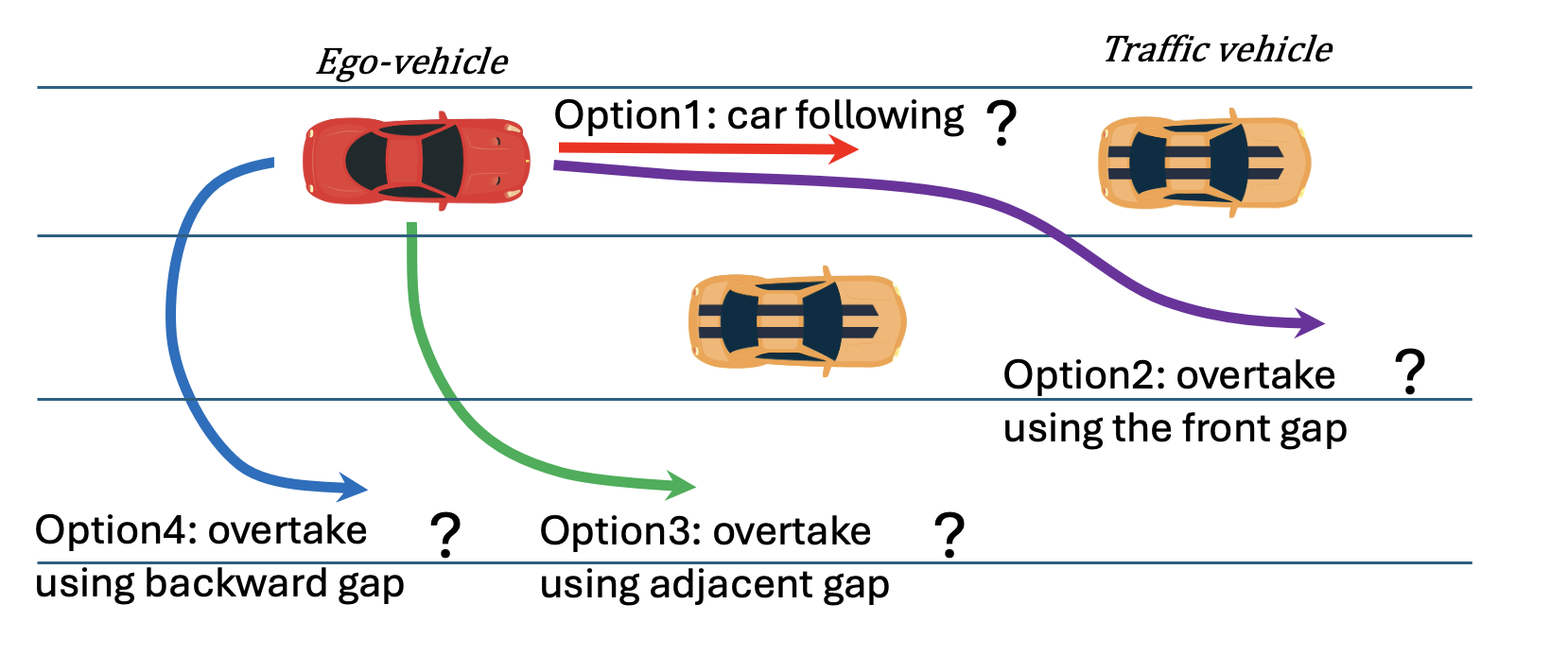}
\caption{In complex traffic conditions, multiple driving strategies may be available, including conservative but simpler behaviors such as maintaining a safe following distance behind the lead vehicle and more aggressive and complicated maneuvers such as exploiting available front gaps to execute overtaking.}  
\label{options}
\end{figure}

As illustrated in Figure~\ref{options}, the overtaking scenario we consider in this study involves two or more slower vehicles ahead of the ego vehicle and offers multiple potential driving strategies, including (1) lane keeping with car-following, (2) overtaking via a forward gap, (3) overtaking via the adjacent gap, and (4) overtaking by a backward gap. The first three options are relatively easy for RL agents to discover as they involve fewer undesirable actions such as decelerations or lane changes.  
In contrast, the fourth option is safe, general, and relatively easy for humans to identify and design but challenging for RL agents to learn because the strategy requires multiple intermediate steps that may yield short-term negative rewards, discouraging exploration toward this option. We adopt the fourth option as the demonstration policy. This safe strategy is crucial for enabling the agent to overtake effectively in complex scenarios and to realize the large long-term expected reward after overtaking. By incorporating demonstrations of this option, we encourage the agent to explore such “undesirable” intermediate steps more frequently and to optimize its overtaking policy accordingly.

Our contribution can be summarized as follows: 
\begin{enumerate}
\item We introduce a multilane highway overtaking scenario designed to test an RL agent’s ability to overcome exploration barriers in pursuit of an optimal strategy.
\item We propose a rule-based suboptimal overtaking controller that demonstrates a feasible approach to handle the highway overtaking scenario.
\item We present a DRL framework that leverages a demonstration controller to bootstrap online RL training and evaluate its ability to overcome exploration barriers in pursuit of an optimal strategy.
\end{enumerate}

The paper is organized as follow: Section~\ref{sec_two} reviews common methods for using prior information to improve RL performance. Section~\ref{sec_three} introduces the problem formulation and the DRL-based controller. Section~\ref{sec_four} describes our proposed RL-based driving framework guided by a suboptimal demonstration controller. Section~\ref{sec_five} presents experimental results in an overtaking scenario, including comparisons with baseline methods. Finally, Section~\ref{sec_six} concludes the paper.

\begin{figure*}[!t]
\centering
\includegraphics[width=1.1\textwidth]{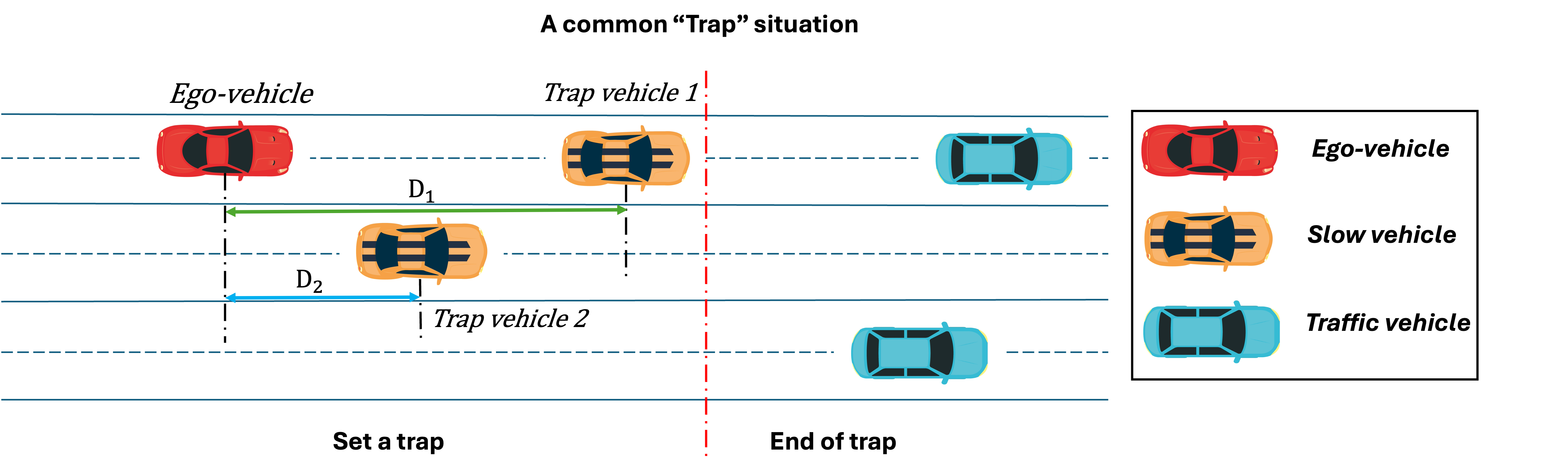}
\caption{A highway "Trap" situation is represented by two vehicles moving slower than the typical traffic, forcing the ego vehicle to reduce speed to avoid a collision. The agent's objective is to maneuver out of this "Trap" formed by slow-moving traffics and subsequently accelerate to the ideal cruising speed.}
\label{explain senario1}
\end{figure*}

\section{Related Work}
\label{sec_two}
Early research on RL-based autonomous driving primarily focused on a driving environment with sparse traffic and predictable traffic patterns to manage complexity~\cite{sonu2018exploiting,wang2020learning,chen2018deep,xu2018reinforcement}. Within these simplified domains, RL agents effectively learn driving policies by repeated interactions with a simulated environment. Although these approaches have demonstrated success in basic tasks, such as lane-keeping and straightforward car-following scenarios, they often encounter significant challenges related to exploration efficiency and sample inefficiency in some critical driving scenarios. Agents frequently settle into suboptimal behaviors, particularly in complex and safety critical scenarios involving delayed reward feedback and long-horizon decision-making.

To address these challenges and enhance sample efficiency, researchers have incorporated prior domain knowledge into RL frameworks for driving. This prior knowledge commonly originates from rule-based or model-based methods, as well as from human expert demonstrations.

Rule-based and model-based methods have been widely explored for providing structured guidance to RL training. One strategy uses rule-based or model-based methods as soft and hard constraints throughout the entire training process~\cite{yurtsever2020integrating, wang2019lane}. Although these methods provide structured guidance and accelerate learning, they often lack flexibility in assessing dynamic traffic conditions and vehicle dynamics. Consequently, RL agents can become overly dependent on fixed rules or waypoint trajectories, severely limiting their ability to adapt to real-time uncertainties or rapidly changing driving environments. Such dependence can cause failures in scenarios that require immediate deviation from planned trajectories due to sudden obstacles or unexpected events. 
Alternatively, some integration methods utilize RL to optimize parameters within rule-based or model-based frameworks~\cite{likmeta2020combining, wang2023combined} or adopt hierarchical designs in which RL is applied to high-level decision-making modules~\cite{lubars2021combining, moghadam2019hierarchical, shi2019driving}. These hybrid strategies offer a balanced trade-off between interoperability and adaptability. However, their effectiveness still heavily depends on the accuracy and reliability of the underlying rule-based or model-based components. Overly restrictive rules might also hinder the discovery of an optimal driving strategy.

Human expert demonstrations represent another critical source of prior knowledge for online RL training. These demonstrations leverage the expertise of experienced human drivers, either through real-time human-in-the-loop interactions or through off-line expert datasets~\cite{wu2022prioritized}. In interactive RL scenarios, a human expert actively provides immediate corrective feedback during training, improving learning efficiency and minimizing undesirable exploratory behavior. However, this approach requires substantial human effort and continuous online supervision, making it labor-intensive and potentially difficult to scale. In contrast, offline learning methods, including imitation learning or offline RL, enable agents to efficiently bootstrap policy training by mimicking expert behaviors extracted from high-quality driving datasets~\cite{han2019driving,bojarski2016end,bojarski2017explaining,codevilla2018end,hawke2020urban}. Although this approach is more scalable in principle, collecting comprehensive, high-quality datasets can be costly.

\begin{figure*}[!t]
\centering
\includegraphics[width=0.9\textwidth]{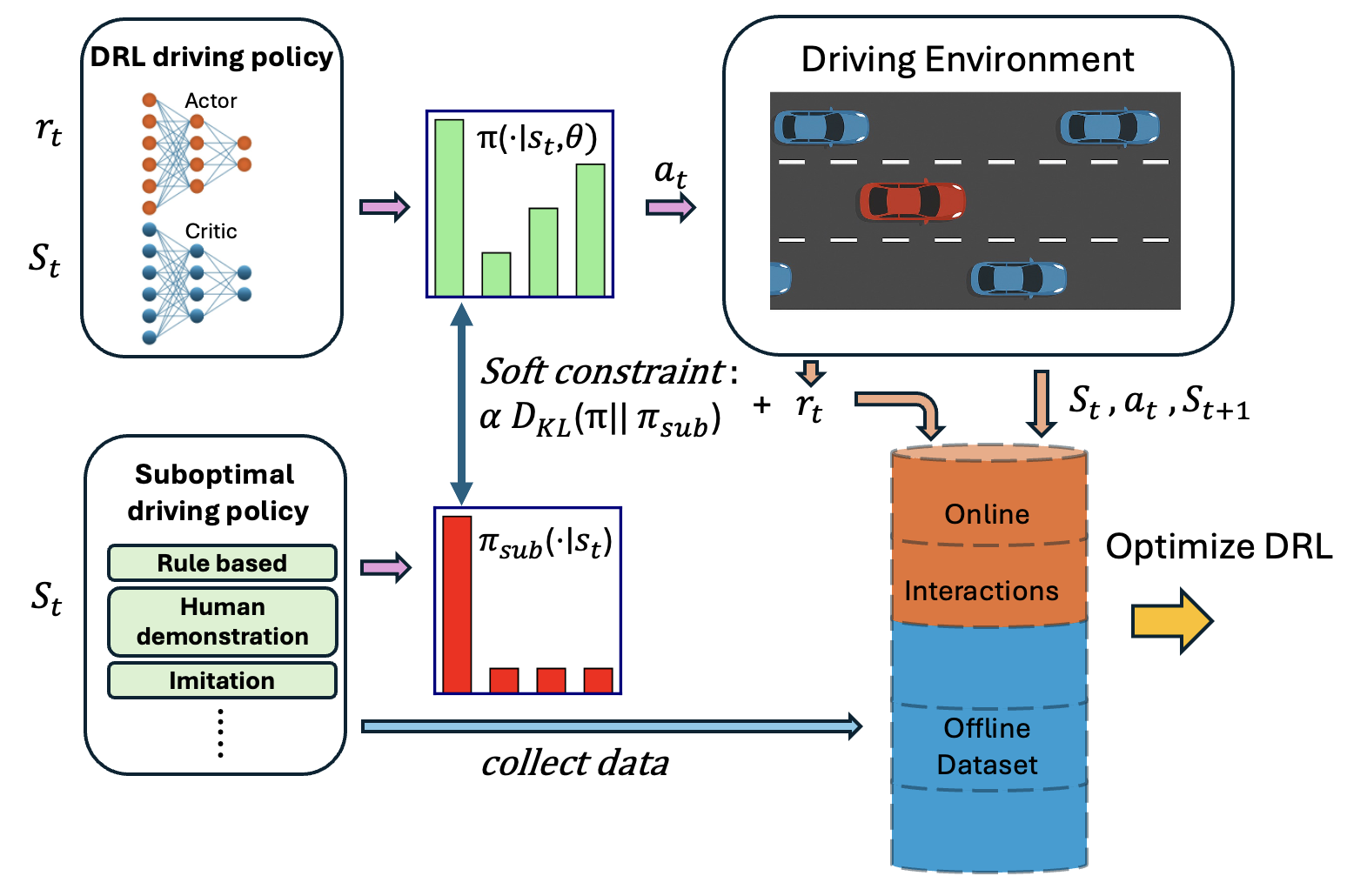}
\caption{We leverage the suboptimal policy in two ways: (1) as a soft constraint on the RL policy during initial training stages, and (2) as a source for populating the replay buffer with additional training samples.}
\label{training framework}
\end{figure*}

We propose a DRL-based autonomous driving framework specifically designed to address critical driving scenarios. Unlike previous approaches that rely heavily on expert demonstrations or optimal control methods, our framework utilizes a suboptimal controller as guidance~\cite{kendall2019learning,huang2022efficient,lu2023imitation,wu2022prioritized}. This suboptimal controller, while not optimal in performance, encapsulates human-like driving heuristics, effectively guiding the RL exploration toward practical driving behaviors. Compared to optimal control designs, developing a heuristic controller significantly reduces design complexity and costs while simultaneously enhancing sample efficiency aligned with human driving strategies. As shown in Figure~\ref{training framework}, we integrate driving heuristics into the RL framework as a soft constraint and as an additional source of training samples. The incorporation of heuristic guidance balances structured insights from human expertise with the necessary flexibility, enabling the RL agent to efficiently explore and ultimately identify better driving policies.

\section{Problem Formulation}
\label{sec_three}
We formulate the highway driving problem as a Markov Decision Process (MDP) 
\begin{equation}
\mathcal{M} \;=\; 
\bigl(\,\mathcal{S},\;\mathcal{A},\;\mathcal{P},\;r,\;\gamma\bigr)
\end{equation}
where $\mathcal{S}$ is the set of states, $\mathcal{A}$ is the set of possible actions, 
$\mathcal{P}$ represents the state transition dynamics, $r:\mathcal{S}\times\mathcal{A} \to \mathbb{R}$ 
is the reward function, and $\gamma \in (0,1)$ is the discount factor. 
Our goal is to learn a policy $\pi(a \mid s)$ that maximizes the expected cumulative discounted reward:
\begin{equation}
\max_{\pi} \;\mathbb{E}\Bigl[\,\sum_{t=0}^{\infty}\gamma^t\,r\bigl(s(t),\;a(t)\bigr)\Bigr], 
\quad 
a(t)\sim\pi(\cdot\mid s(t)).
\end{equation}

\subsection{Critical Driving Scenarios}
\label{Critical driving scenarios}
We utilize a critical driving scenario that is notably challenging for standard DRL agents to find an optimal driving strategy. As illustrated in Figure~\ref{explain senario1}, upon entering the main road, the ego vehicle encounters two slower vehicles ahead, maintaining relatively short distances ${D_1}$ and ${D_2}$. These slower vehicles create a traffic "trap" by driving significantly slower than the surrounding traffic and below the ego vehicle's preferred cruising speed. To escape from the "trap", the ego vehicle needs to execute an overtaking maneuver by utilizing the backward gap. From an RL perspective, successfully performing this overtaking maneuver requires the agent to recognize that the future rewards gained by escaping the slow vehicle trap exceed those obtained by remaining trapped. This potential for increased reward, however, is not immediately apparent until the ego vehicle successfully exits the trap. 

Our simulated traffic scenario consists of two vehicle types: the "trap" vehicles intentionally driving at lower speeds and regular traffic vehicles controlled by the Intelligent Driver Model (IDM)~\cite{treiber2000congested} and Minimizing Overall Braking Induced by Lane changes (MOBIL) controllers~\cite{kesting2007general}. Incorporating additional dynamic traffic vehicles beyond the trap scenario serves several important purposes. First, it prevents the learning agent from overfitting to a fixed traffic pattern. Lastly, the presence of additional traffic ensures that the RL agent must learn both effective overtaking strategies and adaptive maneuvering skills in realistic, complex driving conditions, fostering more robust and generalizable policy learning. Simulation settings during training and testing is identical and can be found in Table~\ref{table:Simulation setting}. Detailed configurations of the traffic vehicles follow our previous work~\cite{zhang2025extensive}.
\begin{table}[b]
\centering
\begin{tabular}{l l l} \hline
 meaning & Value \\ \hline
Vehicle length    &   $5m$  \\
Vehicle width   & $ 2m$  \\
Road width   & $ 4m$  \\ 
Initial longitudinal distance to trap veh 1 ${D_1}$ & [14.80,16.44] m\\ 
Initial longitudinal distance to trap veh 2 ${D_2}$ & [4.06,7.43] m \\\hline
\end{tabular}
\vspace{5mm}
\caption{Simulation settings}
\label{table:Simulation setting}
\vspace{-4mm}
\end{table}
\subsection{Traffic Vehicle Control}

We employ the IDM  for longitudinal car-following dynamics and the MOBIL model for lane-change decisions. IDM calculates the longitudinal acceleration by considering the current velocity of the vehicle, the distance to the lead vehicle, and the relative speed. Specifically, given a gap $s$, a speed $v$, and a speed difference $\Delta v$ with respect to the vehicle ahead, the IDM acceleration $a$ is defined as:
\begin{equation}
\label{eq:idm_acc}
a = a_{\max} \left[ 1 - \left(\frac{v}{v_{\text{desired}}}\right)^\delta 
- \left(\frac{s^*(v, \Delta v)}{s}\right)^2 \right],
\end{equation}
where $a_{\max}$ is the maximum acceleration, $v_{\text{desired}}$ is the desired cruising speed, and $\delta$ is an exponent value. The desired gap $s^*$ captures the minimum distance required to maintain safety and is computed as:
\begin{equation}
\label{eq:idm_desired_gap}
s^*(v, \Delta v) = s_0 + T \, v + \frac{v \, \Delta v}{2 \sqrt{a_{\max} \, b}},
\end{equation}
where $s_0$ is the minimum standstill gap, $T$ is the desired time headway, and $b$ is the comfortable deceleration. This formulation balances free-flow acceleration and collision avoidance within traffic streams.

For lateral movements, the MOBIL model triggers lane changes when the overall benefit, accounting for both the ego vehicle and its neighbors, exceeds a threshold. The incentive criterion is expressed as:
\begin{equation}
\label{eq:mobil_incentive}
\Delta a_{\text{ego}} + p \, (\Delta a_{\text{rear}} + \Delta a_{\text{front}}) \;>\; a_{\text{threshold}},
\end{equation}
where $\Delta a_{\text{ego}}$ is the ego vehicle's acceleration gain, $\Delta a_{\text{rear}}$ and $\Delta a_{\text{front}}$ denote changes in acceleration of the affected trailing and leading vehicles in the target lane, and $p$ is a politeness factor. A safety criterion further ensures no vehicle is forced to brake beyond a comfortable limit. By combining IDM and MOBIL, we capture realistic traffic interactions: vehicles maintain sensible headways and make lane-change decisions that balance individual and collective benefits. Traffic vehicles parameters can be found in Table~\ref{Traffic vehicle parameters}.

\begin{table}[ht]
\centering
\begin{tabular}{l l l } \hline
Symbol & meaning & Value \\ \hline
$a_{\max}$ &Maximum acceleration & $0.5m/s^2$  \\
$v_{\text{desired}}$  &  Desired cruising speed &  $12.5m$\\
 $s_0$ &Desired distance gap&$10m$\\
 $T$&Desired time gap&$1.5s$\\
$b$&Comfortable deceleration&$0.5m/s^2$\\
$a_{\text{th}}$&Acceleration threshold & $0.2m/s^2$\\
$p$&Politeness factor&$0.5$\\
 ${\delta}$ &Exponent value &  $4$ \\\hline
\end{tabular}
\vspace{5mm}
\caption{Traffic vehicle parameters}
\label{Traffic vehicle parameters}
\vspace{-4mm}
\end{table}

\subsection{Deep Reinforcement Learning}
We employ a discrete-action variant of SAC for policy optimization. 
We adapt SAC by representing the policy $\pi(a\mid s)$ as a categorical distribution over 9 possible discrete actions described below. 
The objective then becomes:
\begin{equation}
J(\pi) = \mathbb{E}_{\tau \sim \pi} \left[ \sum_{t=0}^{\infty} \gamma^t \left( r(s_t, a_t) + \alpha \, \mathcal{H} \big( \pi(\cdot \mid s_t) \big) \right) \right]
\end{equation}
where $\mathcal{H}$ denotes the entropy of the policy distribution, 
and $\alpha>0$ is an entropy temperature parameter. 
This formulation preserves SAC's focus on maximizing both the expected return and the policy's entropy, 
yielding robust and exploratory driving policies in the discrete control setting. The RL training hyper-parameters can be found in Table~\ref{table:RLparam}.

\subsubsection{State}

Each state $s\in \mathcal{S}$ encodes both the ego vehicle’s motion and local traffic context as defined by 
\begin{equation}
\begin{split}
s = \Bigl( &\,p^e,\; x^e,\; y^e,\; v^e_{\mathrm{lon}},\; v^e_{\mathrm{lat}},\; d_e, \\
           &\quad \{p^n,\; \Delta x^n,\; \Delta y^n,\; \Delta v_{\mathrm{lon}}^n,\; \Delta v_{\mathrm{lat}}^n\}_{n=1}^{4} \Bigr)
\end{split}
\end{equation}
where $p^e$ denotes the presence of ego vehicle, $x^e$,$y^e$ is the ego vehicle’s longitudinal and lateral position,  $v^e_{\mathrm{lon}},v^e_{\mathrm{lat}}$ is the ego vehicle’s longitudinal and lateral velocity, $d^e$ is its lateral offset from the lane center, 
$p^n\in\{0,1\}$ indicates the presence of the $n$-th traffic vehicle, and  
$\Delta x^n$, $\Delta y^n$, $\Delta v_{\mathrm{lon}}^n$, and $\Delta v_{\mathrm{lat}}^n$ denote the relative position and velocity of the $n$-th surrounding vehicle with respect to the ego vehicle in both the longitudinal and lateral directions.

\subsubsection{Action Space}
We define a discrete action set $\mathcal{A}$ consisting of 9 actions, 
each represented by an acceleration and steering pair $(\dot{v}, \theta)$ chosen from the sets
\begin{equation}
\begin{split}
\dot{v} \;\in\; \{-1,\;0,\;1\}\;\mathrm{m}/\mathrm{s}^2, \\
\theta \;\in\; \Bigl\{-\tfrac{\pi}{50},\;0,\;\tfrac{\pi}{50}\Bigr\}\;\mathrm{rad}.
\end{split}
\end{equation}

\subsubsection{Reward}

The reward function \( r(s,a) \) combines terms for velocity tracking, lane centering, steering smoothness, and accident penalties. The total reward is formulated as follows:
\begin{equation}
r(s,a) = 
\begin{cases} 
\dfrac{w_v r_v + w_y r_y + w_{\theta} r_{\theta}}{w_v + w_y + w_{\theta}}, & \text{no accident} \\[6pt]
-10, & \text{accident}
\end{cases}
\end{equation}
where the weights \( w_v, w_y, w_{\theta} \) balance each reward component's contribution. The value of each weight is show in Table~\ref{table:RLparam}.

The velocity reward \( r_v \) encourages maintaining an optimal speed (\( v^e_{\mathrm{lon}}=15\,\text{m/s} \)) and is defined piecewise to penalize speeds that are either too low or excessively high:
\begin{equation}
r_v = 
\begin{cases} 
\exp\left(-(v^e_{\mathrm{lon}} - 15)^2\right), & v^e_{\mathrm{lon}} > 15 \\[4pt]
\frac{8}{25}v^e_{\mathrm{lon}} - \frac{19}{5}, & 12.5 < v^e_{\mathrm{lon}} \leq 15 \\[4pt]
\frac{2}{75}v^e_{\mathrm{lon}} - \frac{2}{15}, & 5 < v^e_{\mathrm{lon}} \leq 12.5 \\[4pt]
0, & v^e_{\mathrm{lon}} \leq 5. 
\end{cases}
\end{equation}

The lane-centering reward \( r_y \) promotes maintaining a central lane position and is formulated as:
\begin{equation}
r_y = \exp(-1.5\,{d_e}^{2})
\end{equation}
where \( d_e\) represents the vehicle's lateral offset from the lane center.

The steering reward \( r_{\theta} \) penalizes excessive steering angles, thereby encouraging smoother maneuvers:
\begin{equation}
r_{\theta} = -|\sin(\theta)|.
\end{equation}

The accident penalty applies a negative reward upon critical failures, including collision, road departure, or dangerous stoppage, to strongly discourage unsafe behaviors. A detailed explanation of the reward design can be found in our previous study~\cite{zhang2025extensive}.

\subsubsection{Successful Completion}
We define successful completion in terms of the ego vehicle’s ability to overtake the leading slow-moving traffic, referred to as "Trap Vehicle 1" in Figure~\ref{explain senario1}. Specifically, a trial is considered successful if the ego vehicle overtakes Trap Vehicle 1 and remains ahead of it for the remainder of the episode.
If the ego vehicle crashes or departs from the lane after overtaking "Trap Vehicle 1", the episode is still marked as a successful escape. However, such critical failures are accounted for separately under the collision rate metric in the safety evaluation.

\begin{table}[!t]
\centering
\begin{tabular}{l l l } \hline
Symbol & meaning & Value \\ \hline
 $\gamma$ & Discount factor & 0.8 \\
 $\alpha_a$ & Learning rate (actor network) & $5e-4$ \\
 $\alpha_c$ & Learning rate (critic network) & $1e-3$ \\
 $N$ & Batch size  & 64\\
 $D$ & Reply buffer size   & 50,000 \\
$\beta$ & Intial offline ratio   & 0.6 \\
 $\eta$  & Target entropy  & -1 \\
 $\alpha_{init}$& Initial entropy coefficient  &0.01  \\
 $\alpha_{lr}$ & Entropy coefficient learning rate  &$1e-3$  \\
  $w_v$ & Speed reward weight & 1.5 \\
$w_{\theta}$ & Steering reward weight & 0.05\\
$w_y$ & Lane centering reward weight & 0.05 \\
$w_{kl}$ & Initial soft constraint weight & 2 \\
\hline
\end{tabular}
\vspace{5mm}
\caption{RL hyper-parameters}
\label{table:RLparam}
\vspace{-4mm}
\end{table}

\subsection{Rule-based Suboptimal Controller}

We introduce a rule-based suboptimal controller designed specifically for overtaking scenarios, as illustrated in Algorithm~\ref{alg:select_speed_lane}. The controller utilizes an overtaking decision flag $O_{\mathrm{dec}}$, which indicates whether an overtaking maneuver is permitted and is safe. Specifically, the flag is set to true only when the following safety conditions are satisfied:
\begin{itemize}
\item The ego vehicle will not rear-end a slower vehicle in the target lane after the lane change.
\item The ego vehicle will not be rear-ended by a faster vehicle already in the target lane.
\end{itemize}
In our current scenario, there are no following vehicles in either the current or target lane, so the flag is manually set to true to allow the overtaking behavior. If overtaking is enabled ($O_{\mathrm{dec}} = \mathrm{true}$), the controller may change lane to the target lane index $l_{\mathrm{target}}$, otherwise the ego vehicle remains in its current lane $l_{\mathrm{current}}$.

To ensure safe lane-changing decisions, we employ a unified safety-distance metric, denoted as $d_s$, which represents the minimum required separation distance to the preceding vehicle to perform overtaking maneuvers to the target lane. This safety distance is calculated as follows:
\begin{equation}
\begin{aligned}
d_s &= \bigl(v^e_{\mathrm{lon}} - \min\{\,v_{\mathrm{l\_current}},\,v_{\mathrm{l\_target}}\,\}\bigr)\,t_{\mathrm{safe}}\\
    &\quad+ \Delta x_{\mathrm{l\_current}} - \Delta x_{\mathrm{l\_target}}
\end{aligned}
\end{equation}
where $v^e_{\mathrm{lon}}$ represents the speed of the ego vehicle, $v_{\mathrm{l\_current}}$ and $v_{\mathrm{l\_target}}$ denote the speeds of the current preceding vehicles in the current lane and the speed of the new preceding vehicle in the target lane, respectively, $\Delta x_{\mathrm{l\_current}}$ and $\Delta x_{\mathrm{l\_target}}$ denote the longitudinal distance of the current preceding vehicle and the new preceding vehicle with respect to the ego vehicle, $\Delta x_{\mathrm{l\_current}}-\Delta x_{\mathrm{l\_target}}$ indicates the longitudinal distance between the current preceding vehicle and the new preceding vehicle, and $t_{\mathrm{safe}}$ is the safety time distance to the preceding vehicle. 

\begin{algorithm}[!t]
\caption{suboptimal heuristic overtaking algorithm}
\label{alg:select_speed_lane}
\begin{algorithmic}[1]
\REQUIRE Environment state $s(t)$, overtaking decision flag $O_{\mathrm{dec}}$, target lane index $l_{\mathrm{target}}$
\WHILE{$O_{\mathrm{dec}} = \text{true}$}
  \IF{$(d_s-x_{\mathrm{l\_current}})> 0$}
    \STATE $(v, l) \gets \bigl(v_{\mathrm{target}},\,l_{\mathrm{target}}\bigr)$
  \ELSE
    \STATE $(v, l) \gets \bigl(v_{\mathrm{target}},\,l_{\mathrm{current}}\bigr)$
  \ENDIF
\ENDWHILE
\STATE \textbf{Output:} Target speed and lane index $(v, l)$
\end{algorithmic}
\end{algorithm}

The target speed for the ego vehicle, denoted by $v_{\mathrm{target}}$, is determined by selecting the maximum feasible speed from a predefined set $V$, which by default is comprised of $n$ uniformly spaced values that are slower than the minimum speed of both the current preceding vehicle and the new preceding vehicle, that is 
\begin{equation}
\begin{aligned}
v_{target}= \max_{i=1 \ldots n}\Bigl\{\,v_i&\in \bigl\{v_{min}+\frac{i-1}{n-1}(v_{\max}-v_{\min})\bigr\}  
\;\Bigm|\; \\[0ex]
&v_i <\min\bigl(v_{\mathrm{l\_current}},v_{\mathrm{l\_target}})\Bigr\}
\end{aligned}
\end{equation}
with $v_{\min}=6\ \mathrm{m/s}$, $v_{\max}=15\ \mathrm{m/s}$, and $n=5$.
This selection ensures that a safe distance is maintained to all the trap vehicle while performing the overtaking behavior.

%

In conclusion, the controller initiates the overtaking maneuver by identifying the new preceding vehicle in the target lane and maintaining a safe overtaking distance from the current preceding vehicle. The actual lane change is then executed by tracking the designated target speed and lane index, which are subsequently handled by the lower-level rule-based controller~\cite{zhang2025extensive}.

\section{Bootstrapping RL via Suboptimal Policy Framework}
\label{sec_four}
We integrate a suboptimal rule-based controller into the RL framework to enhance learning efficiency and guide exploration. Specifically, we leverage the suboptimal policy in two ways: (1) as a soft constraint on the RL policy during initial training stages, and (2) as a source for populating the replay buffer with additional training samples.

\subsection{Soft Constraint through KL Divergence}
Instead of directly imitating the rule-based policy, we apply a Kullback-Leibler (KL) penalty to nudge the learned policy $\pi$ toward the suboptimal strategy during initial training~\cite{huang2022efficient}. Specifically, we modify the entropy-regularized RL objective as
\begin{equation}
\label{eq:rl_kl_objective_discounted}
\begin{aligned}
\max_{\pi} \;\mathbb{E}_{\tau \sim \pi} \Biggl[ &\sum_{t=0}^{\infty} \gamma^t \Bigl(
\, r(s_t,a_t) \;+\; \alpha \,\mathcal{H}\bigl(\pi(\cdot \mid s_t)\bigr) \\[1ex]
&{}-\; \xi \, D_{\mathrm{KL}}\!\Bigl(\pi(\cdot \mid s_t)\;\|\;\tilde{\pi}^{\text{rule}}(\cdot \mid s_t)\Bigr)
\,\Bigr) \Biggr]
\end{aligned}
\end{equation}
where $\alpha > 0$ trades off reward maximization against policy entropy $\mathcal{H}(\pi)$ and $\xi \ge 0$ controls the strength of the KL term. Here, $\tilde{\pi}^{\text{rule}}$ denotes a \emph{stochastic} version of the rule-based policy as defined below. During early training, a higher $\xi$ ensures the agent’s policy stays near the suboptimal policy, but over time this value will be relaxed so that $\pi$ surpasses the demonstrator’s performance. Specifically, we parameterize $\xi$ as a function of the entropy coefficient $\alpha$, setting $\xi = 200 \cdot \alpha$ to maintain a balance between imitation and exploration throughout training.

Our rule-based policy $\mu^{\text{rule}}(s)$ is inherently deterministic, outputting a single best action from a discrete set of $|\mathcal{A}|=9$. To introduce uncertainty, we construct $\tilde{\pi}^{\text{rule}}(\cdot\mid s)$ by assigning a high probability to the deterministic output and small probabilities to other actions. Specifically, if $a_{\text{rule}}$ is the chosen action, we let
\begin{equation}
\label{eq:stoch_discrete_rule}
\begin{aligned}
\tilde{\pi}^{\text{rule}}(a \mid s)
&=
\begin{cases}
p, & \text{if } a = \mu^{\text{rule}}(s), \\[3pt]
\frac{1-p}{|\mathcal{A}|-1}, & \text{otherwise},
\end{cases}\\[4pt]
\end{aligned}
\end{equation}

In practice, we set $p=0.9$, giving the demonstration policy a high chance of selecting the good rule-based action. Introducing stochasticity serves two purposes: First, it aligns action distributions: since SAC policies are inherently stochastic, introducing stochasticity into the suboptimal controller yields a more compatible distribution. This alignment makes KL-based regularization both meaningful and numerically stable. Second, it mitigates over-reliance on the suboptimal controller and promotes robust exploration. By incorporating stochasticity, the agent avoids overfitting to the demonstration policy and is exposed to a broader range of action options. This exposure enables the agent to generalize more effectively and discover improved policies beyond the limitations of the initial suboptimal demonstrations.

\subsection{Integration of Suboptimal Demonstrations}

To guide the agent toward heuristic driving strategies and provide safe reference actions during the early stages of learning, we enrich the replay buffer with demonstration data generated by the suboptimal rule-based controller 
$\tilde{\pi}^{\text{rule}}$. 
We collected 200 episodes of demonstration transitions $(s, a_{\text{rule}}, r(s,a_{\text{rule}}), s')$ by running $\tilde{\pi}^{\text{rule}}$ in the simulated environment. As illustrated in Figure~\ref{training framework}, during training both offline demonstrations and online experience are sampled according to an annealed ratio $\beta$, which decays from 0.6 to 0 over the first 1000 episodes. This ensures early guidance without constraining policy improvement in the long term.

We evaluate two strategies for integrating these demonstrations into learning.

\subsubsection{Q-Value Regularization via Action Margins}

This strategy explicitly regularizes the Q-values by enforcing a margin between the demonstrator’s action and all others~\cite{piot2014boosted,hester2018deep,ibarz2018reward}. The margin supervised loss for each demonstration $(s, a_E)$ is:
\begin{equation}
\label{eq:large_margin_loss}
J_E(Q) = \max_{a \in \mathcal{A}} \left[ Q(s,a) + l(a_E,a) \right] - Q(s,a_E),
\end{equation}
where $l(a_E, a)$ is a margin function defined as:
\begin{equation}
\label{eq:margin_function}
l(a_E, a) = \begin{cases}
0, & \text{if } a = a_E, \\[5pt]
1, & \text{otherwise}.
\end{cases}
\end{equation}

We apply a Q-filter to prevent the agent from strictly imitating suboptimal actions :
\begin{equation}
\label{eq:q_filtered_loss}
J_E^Q(Q) = \begin{cases}
J_E(Q), & \text{if } Q(s,a_E) \ge \max\limits_{a \in \mathcal{A}} Q(s,a), \\[5pt]
0, & \text{otherwise}.
\end{cases}
\end{equation}
This filter ensures that supervision is only applied when the demonstrator's action is at least as good as the agent’s current estimate.

\subsubsection{Demonstration-Guided Reward Augmentation}

As an alternative to explicit supervision, we implement reward shaping by adding a bonus to demonstration actions~\cite{reddy2019sqil}
\begin{equation}
    r'(s,a_E) = r(s,a_E) + c.
\end{equation}
where $c=2$ is a constant value.
This method implicitly increases the Q-values of demonstrated actions and produces effects similar to large-margin regularization. Empirical study shows that reward shaping yields better long-term performance and generalization, especially when the goal is to surpass or recover from imperfect demonstrations~\cite{reddy2019sqil}.

\section{Experiment}
As described in Section~\ref{Critical driving scenarios}, the simulation is initialized with two "trap" vehicles, followed by traffic vehicles governed by IDM and MOBIL models. The environment is implemented using the \texttt{highway-env} platform~\cite{highway-env}. Simulation settings for both training and testing are identical, and the simulation runs at 2\,Hz. All experiments are conducted on an Ubuntu system equipped with an NVIDIA GeForce RTX~3070\,Ti GPU.

\label{sec_five}
\subsection{Abalation Study}

\begin{figure*}[!t]
  \centering
  \subfloat[]{%
    \includegraphics[width=3.2in]{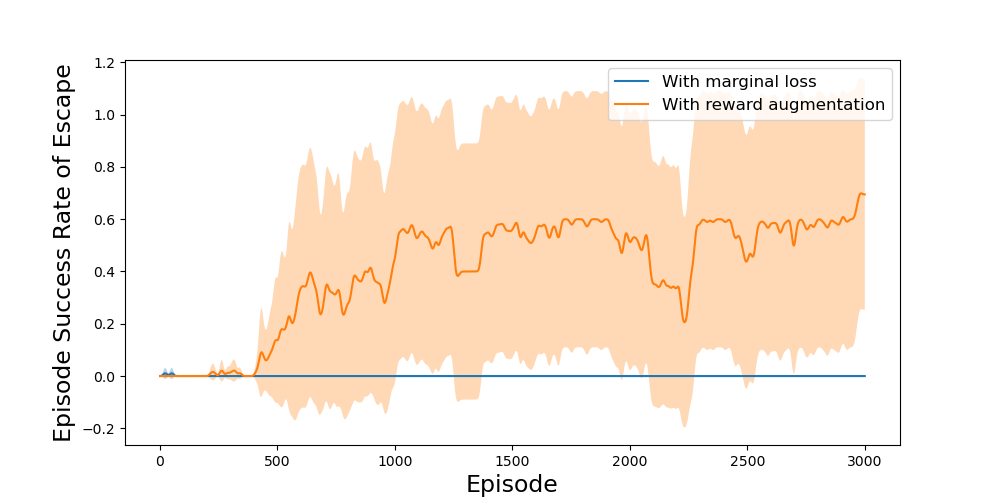}%
    \label{marginal_trap}%
  }\hfil
  \subfloat[]{%
    \includegraphics[width=3.2in]{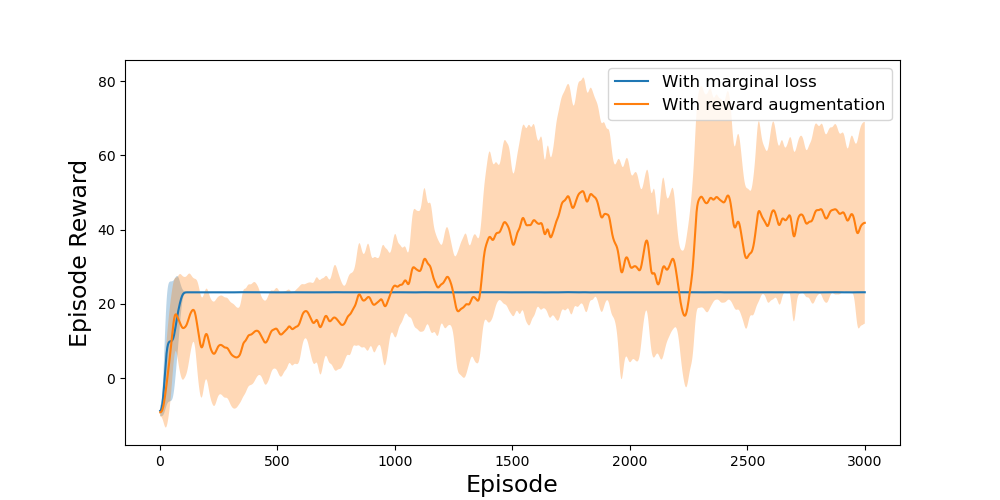}%
    \label{marginal_Reward}%
  }\\
  \vspace{-0.1in}
  \subfloat[]{%
    \includegraphics[width=3.2in]{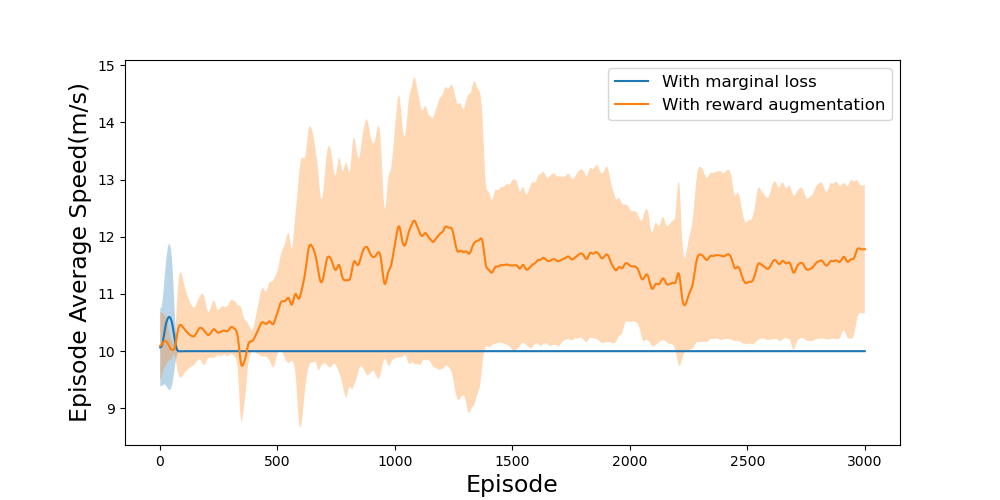}%
    \label{marginal_speed}%
  }\hfil
  \subfloat[]{%
    \includegraphics[width=3.2in]{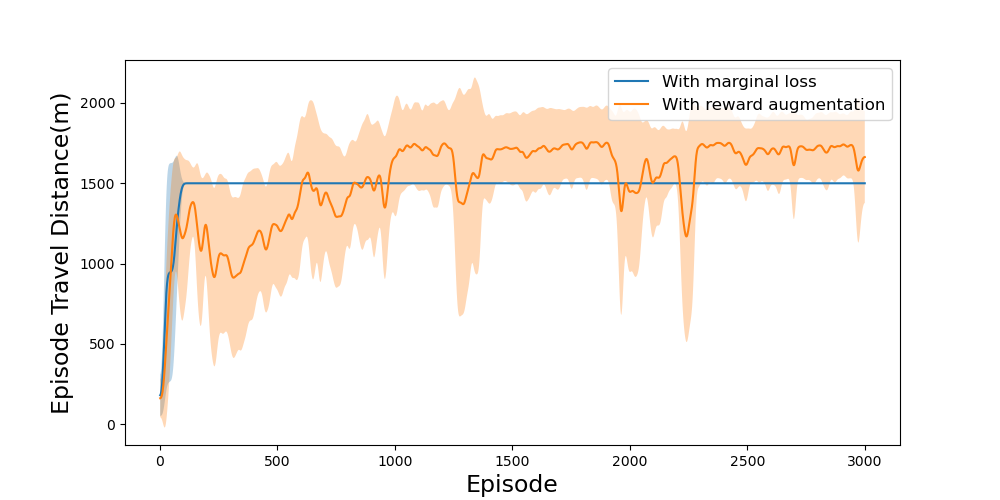}%
    \label{marginal_distance}%
  }
  \caption{Training comparison between integrating demonstration data via marginal loss and reward augmentation: 
    (a) Average success rate in escaping low‑speed traffic; 
    (b) Average reward per episode; 
    (c) Average speed per episode; 
    (d) Average travel distance per episode.
  }
  \label{marginal_training}
\end{figure*}

\begin{figure*}[t]
  \centering
  \subfloat[]{%
    \includegraphics[width=3.2in]{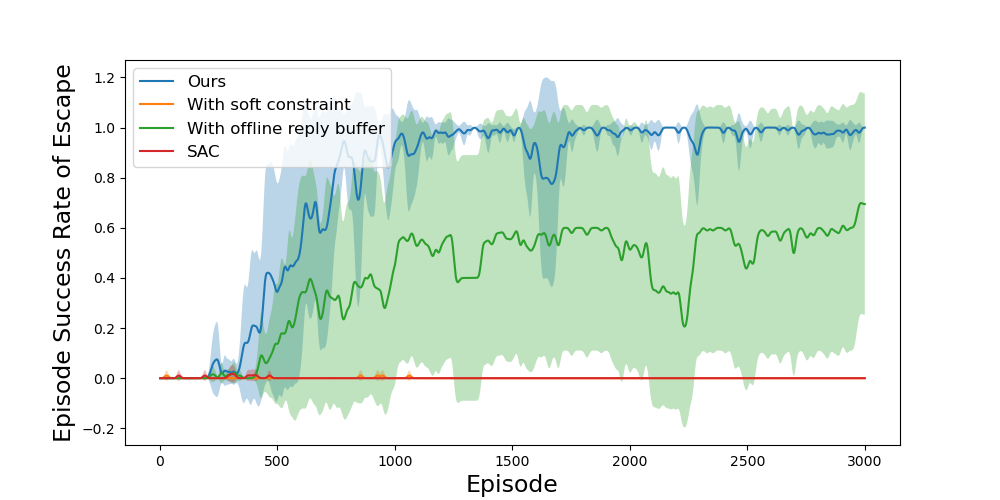}%
    \label{abalationtrap}%
  }\hfil
  \subfloat[]{%
    \includegraphics[width=3.2in]{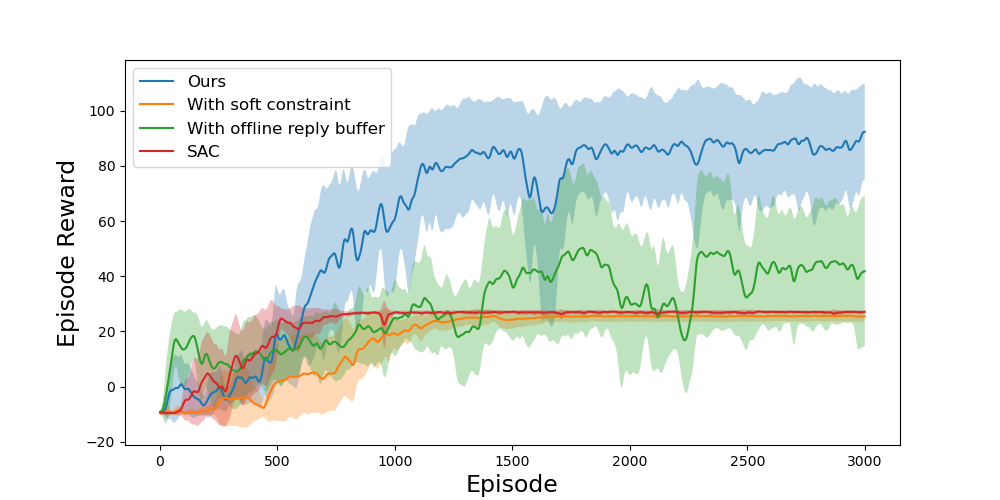}%
    \label{abalationReward}%
  }\\
  \vspace{-0.1in}
  \subfloat[]{%
    \includegraphics[width=3.2in]{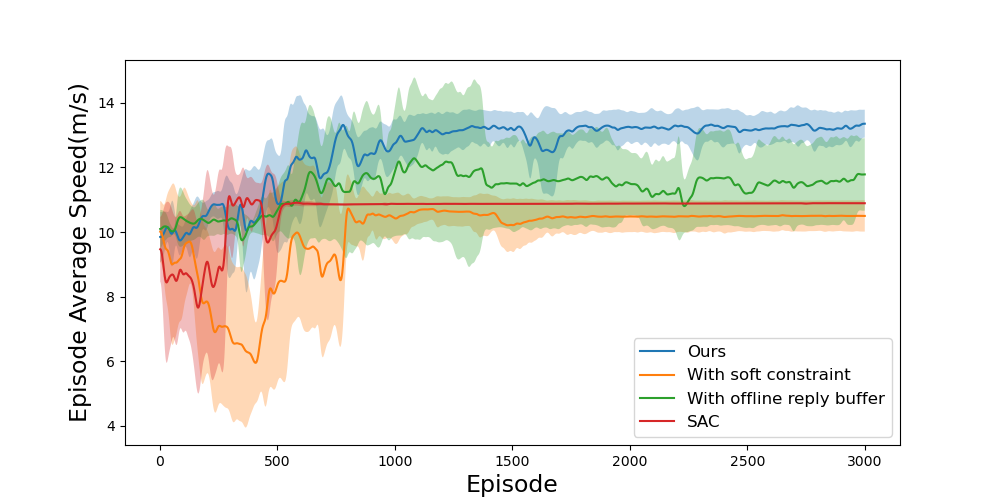}%
    \label{abalationspeed}%
  }\hfil
  \subfloat[]{%
    \includegraphics[width=3.2in]{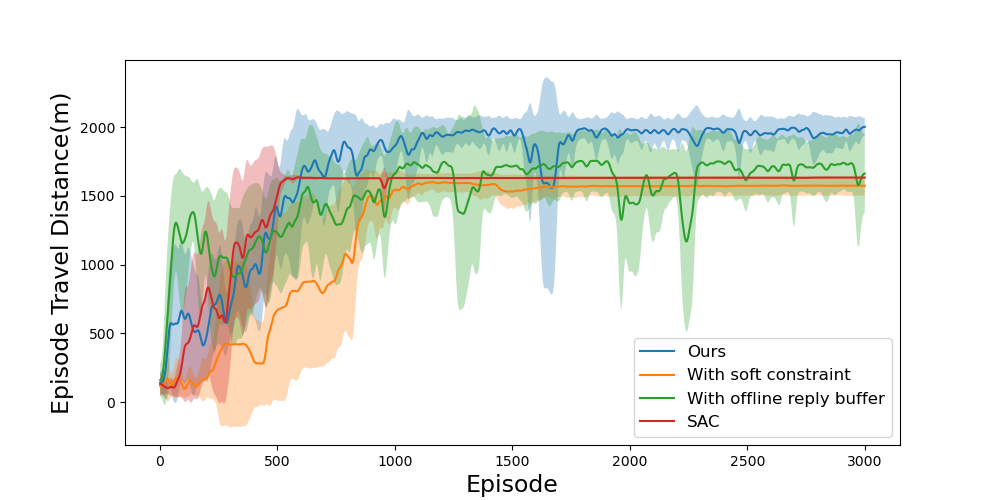}%
    \label{abalationdistance}%
  }
  \caption{Performance during training episodes for SAC with soft constraint, SAC with offline replay buffer, and our method: 
    (a) Average success rate in escaping low‑speed traffic; 
    (b) Average reward per episode; 
    (c) Average speed per episode; 
    (d) Average travel distance per episode.
  }
  \label{abalationtraining}
\end{figure*}

As previously introduced, we integrate the suboptimal driving policy by applying soft constraints and integrating demonstration dataset. 

To find the better strategy for integrating suboptimal demonstration data into RL training, we conduct an ablation study comparing the two approaches. As shown in Figure~\ref{marginal_training}, using reward augmentation leads to better learning outcomes across all metrics compare to margin loss. These results suggest that reward augmentation offers more effective guidance for suboptimal demonstration integration. 

Figure~\ref{abalationtraining} and Table~\ref{table:testing_priors} detail an ablation study examining each approach during training and testing. They demonstrate that adding a reward-augmented offline replay buffer improves the success rate in escaping the trap vehicle scenario, especially during initial training phases, likely due to the diversity in action distributions introduced early on. However, using the reward-augmented offline replay buffer alone introduces training instability and increases the risk of potential collisions. 
Employing SAC with only soft constraints results in slower learning and reduced performance compared to vanilla SAC. This outcome may occur because soft constraints limit SAC's actions to those recommended by the suboptimal policy, reducing exploration potential and adaptability.

The combination of soft constraints with an offline replay buffer appears to better balance the exploration and exploitation trade-off. Soft constraints guide SAC toward safer behaviors initially, while the offline replay buffer provides diverse experiences, thus enabling more robust policy improvement and stable convergence to optimal performance.

\subsection{Performance Comparison}

\vspace{1mm}
\begin{figure*}[!t]
  \centering
  \subfloat[]{%
    \includegraphics[width=3.2in]{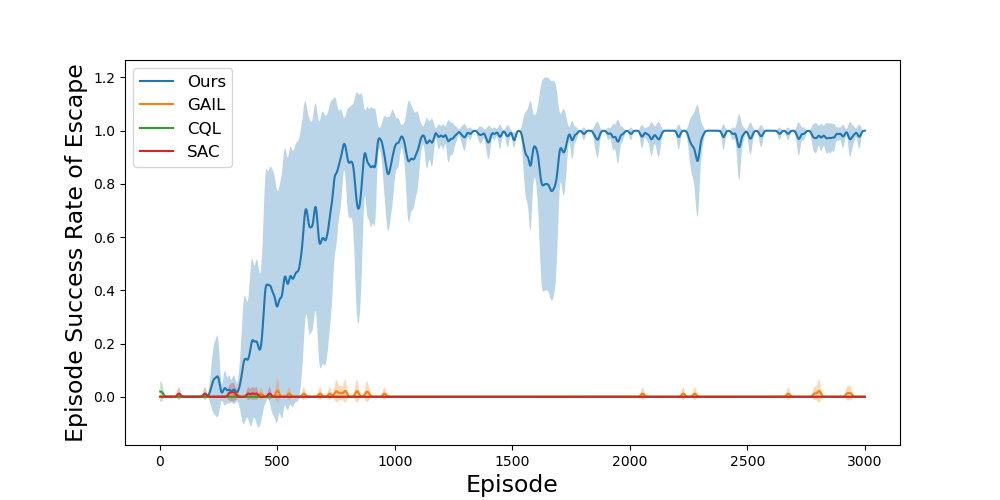}%
    \label{baselinetrap}%
  }\hfil
  \subfloat[]{%
    \includegraphics[width=3.2in]{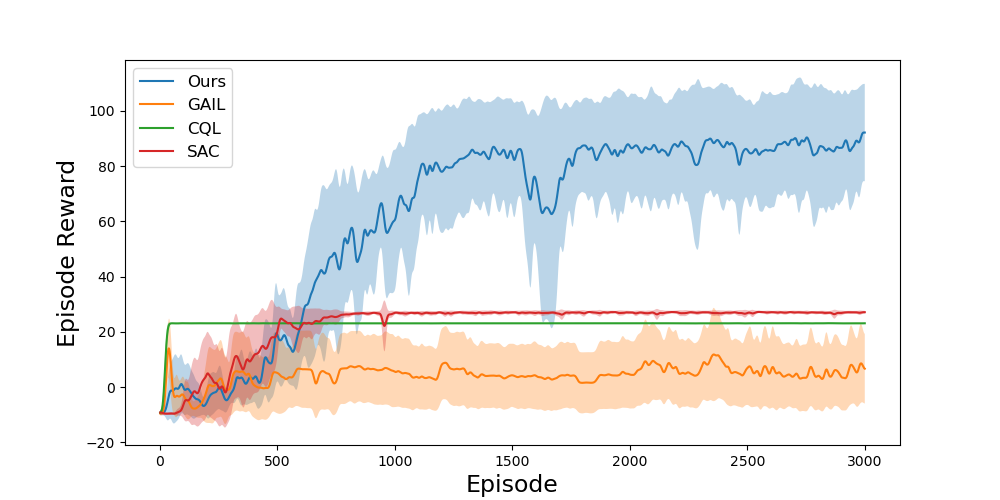}%
    \label{baselineReward}%
  }\\
  \vspace{-0.1in}
  \subfloat[]{%
    \includegraphics[width=3.2in]{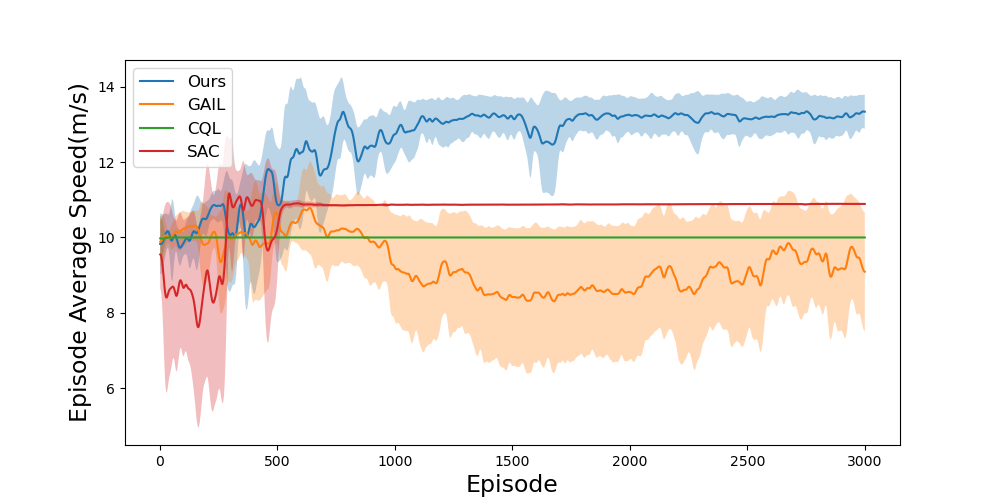}%
    \label{baselinespeed}%
  }\hfil
  \subfloat[]{%
    \includegraphics[width=3.2in]{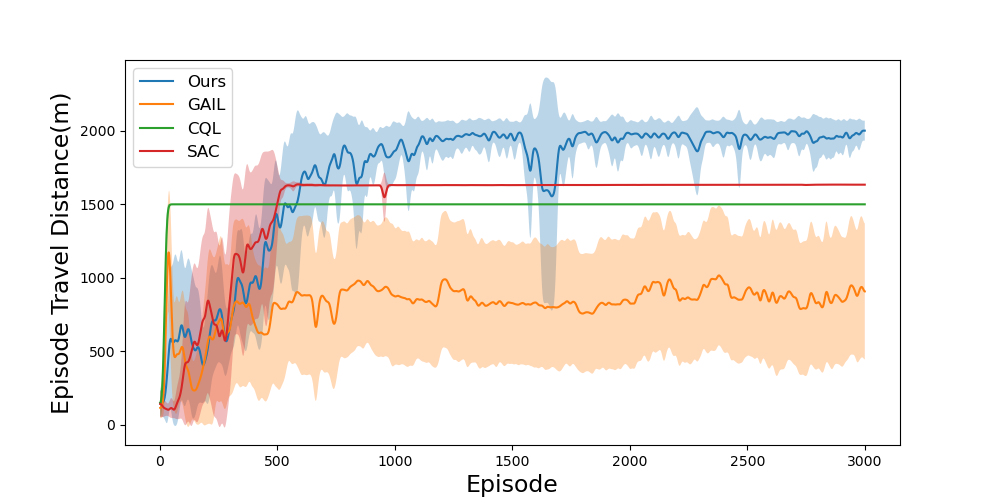}%
    \label{baselinedistance}%
  }
  \caption{Training performance comparison with other RL baseline methods: 
    (a) Average success rate in escaping low‑speed traffic; 
    (b) Average reward per episode; 
    (c) Average speed per episode; 
    (d) Average travel distance per episode.
  }
  \label{baselinetraining}
\end{figure*}

\begin{table*}[!h]
\centering
\resizebox{0.8\textwidth}{!}{
\begin{tabular}{l c c c c c}
\hline
Method & \thead{Success\\rate [\%]} & \thead{Accumulated\\Reward} & \thead{Average\\Speed} & \thead{Travel\\Distance [m]} & \thead{Collision\\rate [\%]} \\
\hline
SAC  & $0 \pm 0$ 
     & $21.32 \pm 7.39$ 
     & $10.50 \pm 1.28$ 
     & $1402.16 \pm 182.87$ 
     & $0 \pm 0$ \\[5pt]
with soft constraint      
    & $0 \pm 0$ 
    & $26.77 \pm 0.94$ 
    & $10.87 \pm 0.02$ 
    & $1631.48 \pm 3.40$ 
    & $0 \pm 0$ \\[5pt]
with offline replay buffer 
    & $100 \pm 0$ 
    & $36.68 \pm 9.66$ 
    & $12.05 \pm 0.39$ 
    & $1804.35 \pm 55.75$ 
    & $5\pm 29.69$ \\[5pt]
\textbf{Ours} 
    & $\mathbf{100 \pm 0}$ 
    & $\mathbf{44.92 \pm 12.62}$ 
    & $\mathbf{12.55 \pm 0.63}$ 
    & $\mathbf{1853.88 \pm 95.46}$ 
    & $\mathbf{0 \pm 0}$ \\
\hline
\end{tabular}%
}
\caption{Comparison results for different methods of integrating suboptimal demonstrations}
\label{table:testing_priors}
\end{table*}

To comprehensively evaluate our proposed approach, we select baseline methods representing RL paradigms and one state-of-the-art behavior cloning method.

\newsec{Conservative Q-Learning(CQL):} CQL~\cite{kumar2020conservative} is an offline RL algorithm designed for learning robust Q-functions from pre-collected datasets without online interaction. By conservatively estimating Q-values, CQL ensures stable training and mitigates overestimation errors common in offline RL settings.


\newsec{SAC:} SAC~\cite{haarnoja2018soft} is an off-policy actor-critic algorithm that optimizes a stochastic policy while maximizing both expected returns and policy entropy. In this paper, SAC serves as an off-policy baseline to illustrate the benefits of bootstrapping online RL with prior demonstration data.
 
\newsec{Generative Adversarial Imitation Learning (GAIL):}GAIL~\cite{ho2016generative} leverages adversarial training to learn policies that mimic expert behaviors from demonstrations without explicitly defining reward functions. GAIL utilizes a discriminator to distinguish between expert and learned behaviors, guiding the policy toward human-like actions.

As shown in Figure~\ref{baselinetraining}, we evaluate the performance of several state-of-the-art online and offline RL methods alongside a behavior cloning method during training. Table~\ref{table:testing_baselines} compares our proposed methods with several baselines, presenting the testing performance of these baseline methods along with the suboptimal controller. 

As Table~\ref{table:testing_baselines} indicates, the suboptimal controller consistently escapes the trap vehicle scenario, confirming that every trajectory in the demonstration dataset successfully escapes the trap. The demonstration dataset used for GAIL and CQL was collected using this suboptimal controller. 

\begin{figure}[!t]
  \centering
  \subfloat[]{%
    \includegraphics[width=0.5\textwidth]{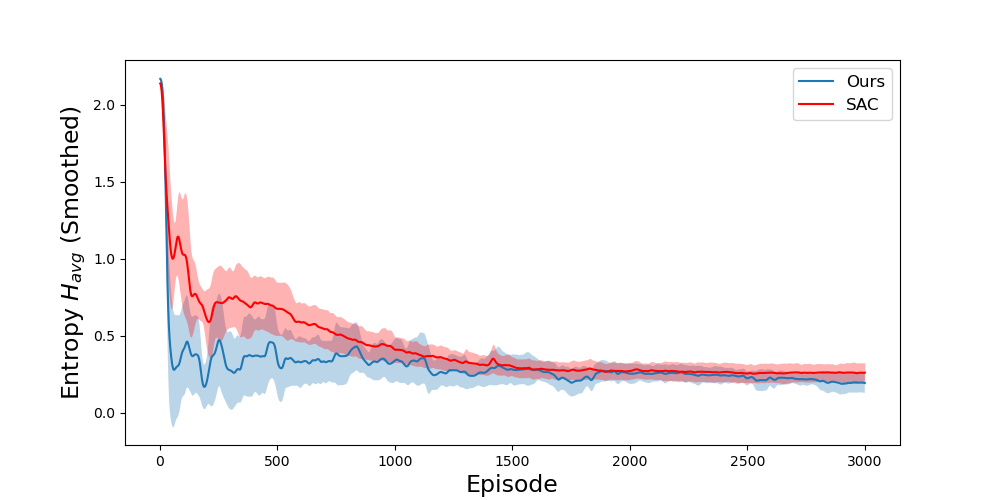}%
    \label{entropy}}
  \\
  \subfloat[]{%
    \includegraphics[width=0.5\textwidth]{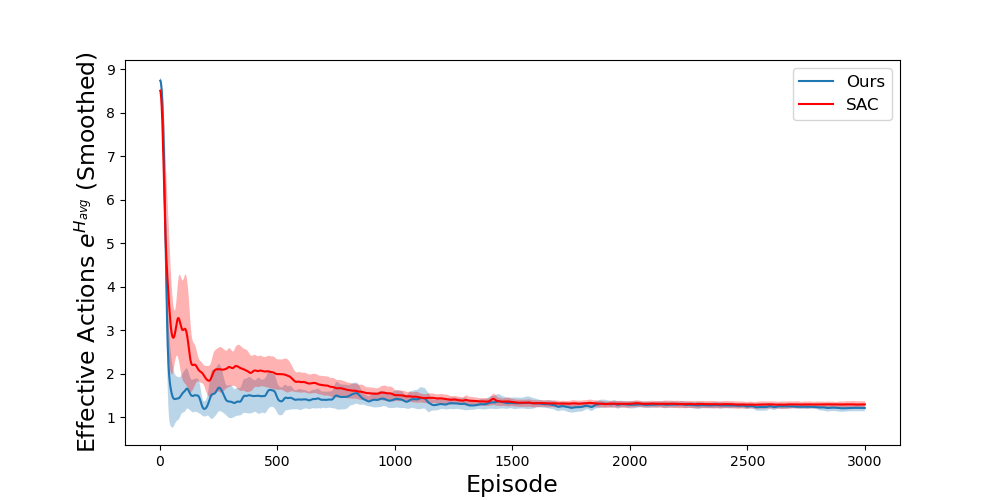}%
    \label{exp_entropy}}
  \caption{Comparison of (a) policy entropy, (b) effective action count during training.}
  \label{fig:entropy_comparison}
\end{figure}
\begin{table*}[!ht]
\centering
\resizebox{0.8\textwidth}{!}{
\begin{tabular}{l c c c c c}
\hline
Method & \thead{Success\\rate [\%]} & \thead{Accumulated\\Reward} & \thead{Average\\Speed} & \thead{Travel\\Distance [m]} & \thead{Collision\\rate [\%]} \\
\hline
CQL  & $0 \pm 0$ 
     & $23.06 \pm 0.15$ 
     & $10 \pm 0$ 
     & $1500 \pm 0$ 
     & $0 \pm 0$ \\[5pt]
GAIL & $0 \pm 0$ 
     & $22.40 \pm 3.31$ 
     & $9.99 \pm 0.02$ 
     & $1488.18 \pm 52.83$ 
     & $5 \pm 22.36$ \\[5pt]
SAC  & $0 \pm 0$ 
     & $21.32 \pm 7.39$ 
     & $10.50 \pm 1.28$ 
     & $1402.16 \pm 182.87$ 
     & $0 \pm 0$ \\[5pt]
Suboptimal Rule                      
    & $100 \pm 0$ 
    & $24.70 \pm 0.44$ 
    & $11.51 \pm 0.8$ 
    & $1727.01 \pm 8.3$ 
    & $0 \pm 0$ \\[5pt]
\textbf{ours} 
    & $\mathbf{100 \pm 0}$ 
    & $\mathbf{44.92 \pm 12.62}$ 
    & $\mathbf{12.55 \pm 0.63}$ 
    & $\mathbf{1853.88 \pm 95.46}$ 
    & $\mathbf{0 \pm 0}$ \\
\hline
\end{tabular}%
}
\caption{Testing comparison with baseline methods}
\label{table:testing_baselines}
\end{table*}

The training and testing results demonstrate consistent patterns across several metrics. The sub-optimal controller yields moderate reward and speed due to its conservative heuristics. 
CQL, GAIL, and SAC fail to achieve meaningful success rates and result in limited accumulated rewards, average speeds, and travel distances. 
They converge to low-risk, low-reward strategies due to unsuccessful exploration in the early stages of training. These methods fail to escape from local optima and instead learn conservative driving behaviors that avoid collisions by maintaining safe following distances and persistently trailing slower vehicles. Such behavior is easy to discover in oversimplified traffic patterns formed by slow-moving “trap” vehicles, where the lack of complex interactions diminishes the need for deeper exploration. As a result, these methods are unable to complete the driving task and consistently achieve 0\% success rates.

Although GAIL and CQL utilize demonstration data effectively, achieving high success rate in replicating demonstrated behaviors, neither outperforms the suboptimal baseline controller in terms of accumulated reward or successful escape rate. During training, GAIL shows the worst performance among all baseline methods. One plausible explanation is that GAIL, being an imitation learning approach, depends heavily on precise discriminator feedback. Given that the demonstration dataset is generated by a suboptimal rule-based controller with limited stochasticity and low behavioral variance, GAIL struggles to generalize beyond the observed trajectories. As a result, it performs worse than methods such as CQL and SAC, which are better equipped to explore and adapt beyond the constraints of the demonstration data.

CQL initially demonstrates the fastest improvement in episodic rewards, benefiting significantly from its offline RL formulation and effective use of demonstration data. Nevertheless, CQL still fails to escape the trap scenario. This limitation arises from its overly conservative optimization strategy, which strongly discourages exploration outside the demonstration distribution, thereby hindering the development of necessary exploratory behaviors crucial for overtaking the trap vehicle.

While SAC outperforms CQL and GAIL in overall performance, it fails to escape the traffic traps. We compare the exploration performance of our proposed method against standard SAC by measuring the average policy entropy and the average effective action count over the course of training. Policy entropy $H_{\mathrm{avg}}$ quantifies the randomness or diversity in the policy's action distribution and is computed from
\begin{equation}
H_{\mathrm{avg}} = -\frac{1}{T}\sum_{t=1}^{T}\sum_{a\in\mathcal{A}}\pi_t(a)\log\pi_t(a),
\end{equation}
where $T$ is the episode length and $\pi_t(a)$ represents the actor’s stochastic distribution over the discrete action set $\mathcal{A}$. Under a uniform policy $\pi_t(a)=1/|\mathcal{A}|$, entropy attains its maximum, indicating maximal exploration.

To provide a more intuitive measure of exploration, we calculate the effective number of actions explored per episode by exponentiating the entropy:
\begin{equation}
\text{Effective actions} = e^{H_{\mathrm{avg}}}.
\end{equation}

Figure~\ref{fig:entropy_comparison} display the exploration dynamics of the proposed method versus vanilla SAC. Both algorithms begin with near‑uniform action distributions, indicating exhaustive initial exploration. SAC maintains higher entropy levels and a larger effective action count longer, reflecting undirected sampling across the entire action set. In contrast, our method rapidly reduces entropy. This rapid reduction shows that the exploration budget is re‑allocated more efficiently and the embedded sub‑optimal controller is able to steer the agent in more effective directions toward promising actions.


\section{Conclusion}
\label{sec_six}

In this paper, we introduced a novel DRL-based autonomous driving framework specifically tailored to address a critical driving scenario. As opposed to traditional methods that rely on optimal control or large-scale expert demonstrations, our approach integrates a suboptimal controller that embeds human-like driving heuristics, effectively guiding the policy toward practical and efficient driving behaviors. The suboptimal controller is employed both as a soft constraint during the early stages of policy training and as an additional source of training samples. Experimental results consistently demonstrated that our framework outperforms standard RL algorithms. These findings highlight the practical benefits of incorporating structured, heuristic-based guidance to accelerate policy training, improve sample efficiency, and enhance autonomous driving performance.

\bibliographystyle{IEEEtran}
\bibliography{ref}
\end{document}